\documentclass[conference]{IEEEtran}
\IEEEoverridecommandlockouts
\usepackage{cite}
\usepackage{amsmath,amssymb,amsfonts}
\usepackage{algorithm}
\usepackage{algorithmic}
\usepackage{graphicx}
\usepackage{textcomp}
\usepackage{xcolor}
\usepackage{booktabs}
\makeatletter
\newcommand{\linebreakand}{%
  \end{@IEEEauthorhalign}
  \hfill\mbox{}\par
  \mbox{}\hfill\begin{@IEEEauthorhalign}
}
\makeatother

\begin{document}

\title{Hetero-SplitEE: Split Learning of Neural Networks with Early Exits for Heterogeneous IoT Devices
\thanks{© 2025 IEEE.  Personal use of this material is permitted.  Permission from IEEE must be obtained for all other uses, in any current or future media, including reprinting/republishing this material for advertising or promotional purposes, creating new collective works, for resale or redistribution to servers or lists, or reuse of any copyrighted component of this work in other works.}}

\author{\IEEEauthorblockN{Yuki Oda}
\IEEEauthorblockA{
\textit{The University of Tokyo}\\
Tokyo, Japan \\
oda-yuki890@g.ecc.u-tokyo.ac.jp}
\and
\IEEEauthorblockN{Yuta Ono}
\IEEEauthorblockA{
\textit{The University of Tokyo}\\
Tokyo, Japan \\
ono-yuta116@g.ecc.u-tokyo.ac.jp}
\linebreakand
\IEEEauthorblockN{Hiroshi Nakamura}
\IEEEauthorblockA{
\textit{The University of Tokyo}\\
Tokyo, Japan \\
nakamura@hal.ipc.i.u-tokyo.ac.jp}
\and
\IEEEauthorblockN{Hideki Takase}
\IEEEauthorblockA{
\textit{The University of Tokyo}\\
Tokyo, Japan \\
takasehideki@hal.ipc.i.u-tokyo.ac.jp}
}

\maketitle

\begin{abstract}
The continuous scaling of deep neural networks has fundamentally transformed machine learning, with larger models demonstrating improved performance across diverse tasks. This growth in model size has dramatically increased the computational resources required for the training process. Consequently, distributed approaches, such as Federated Learning and Split Learning, have become essential paradigms for scalable deployment. However, existing Split Learning approaches assume client homogeneity and uniform split points across all participants. This critically limits their applicability to real-world IoT systems where devices exhibit heterogeneity in computational resources. To address this limitation, this paper proposes Hetero-SplitEE, a novel method that enables heterogeneous IoT devices to train a shared deep neural network in parallel collaboratively. By integrating heterogeneous early exits into hierarchical training, our approach allows each client to select distinct split points (cut layers) tailored to its computational capacity. In addition, we propose two cooperative training strategies, the Sequential strategy and the Averaging strategy, to facilitate this collaboration among clients with different split points. The Sequential strategy trains clients sequentially with a shared server model to reduce computational overhead. The Averaging strategy enables parallel client training with periodic cross-layer aggregation. Extensive experiments on CIFAR-10, CIFAR-100, and STL-10 datasets using ResNet-18 demonstrate that our method maintains competitive accuracy while efficiently supporting diverse computational constraints, enabling practical deployment of collaborative deep learning in heterogeneous IoT ecosystems.\footnote{Accepted at MCSoC 2025}
\end{abstract}


\section{Introduction}

The continuous scaling of deep neural networks has fundamentally transformed the landscape of machine learning. As the number of parameters in machine learning models increases, they have consistently demonstrated improved performance across a wide range of tasks, including computer vision and natural language processing. However, this trend toward larger architectures has introduced significant computational challenges. Training and deploying such models on a single machine increasingly face resource bottlenecks related to processing power, memory capacity, and energy budgets. Consequently, distributed processing has gained attention as an approach to address these scalability challenges, enabling model training and inference to be distributed across multiple machines or devices.

In this context, distributed machine learning frameworks such as Federated Learning (FL)~\cite{McMahan2017-pc} and Split Learning (SL)~\cite{Otkrist2018-av, Vepakomma2018-zg} have attracted renewed interest. These paradigms enable multiple clients to collaboratively train models while preserving data privacy and reducing communication overhead. However, existing distributed learning approaches face fundamental limitations when deployed in real-world Internet of Things (IoT) ecosystems, where devices exhibit heterogeneity in computational resources, including processing power, memory capacity, and energy budgets. In general, FL and SL assume client homogeneity: FL requires all clients to train identical model architectures, while SL imposes uniform split points across all clients. This assumption of homogeneity is often violated in practical IoT scenarios, where resource-constrained sensors, mid-tier edge devices, and powerful edge computers must coexist in the same collaborative learning environment. To accommodate such device heterogeneity, it is desirable to allow each client to use a network size tailored to its computational capacity. However, a naive extension of SL to support heterogeneous clients would impose a high computational load on the server, which must maintain and train separate models for each client, while also suffering from low parallelism as the server becomes a sequential bottleneck. Consequently, it becomes difficult to fully utilize all available computational resources simultaneously across the distributed system.

To address these challenges, we propose \textit{Hetero-SplitEE}. This novel distributed training methodology enables multiple clients with different computational capabilities to train a shared deep neural network in parallel collaboratively.
This approach extends recent work on hierarchical training with early exits~\cite{Sepehri2023-ip}, which primarily focuses on single-client scenarios, to multi-client heterogeneous collaborative training with efficient parallel execution.
Our key innovation lies in allowing each client to select its own split point (cut layer) within the same base architecture.
This flexibility allows resource-constrained devices to train only the initial layers while more powerful devices can train deeper sub-networks, all contributing to the same global model.
Hetero-SplitEE maintains client training entirely locally and does not impose additional alignment; this design choice avoids extra server-to-client communication, keeping the method compatible with bandwidth-constrained IoT deployments.
We further introduce two cooperative training strategies—Sequential and Averaging—that address the server load and parallelism challenges through different trade-offs. The Sequential strategy employs a shared server model to minimize computational overhead, while the Averaging strategy facilitates parallel execution through client-specific server models with periodic cross-layer aggregation.

The main contributions of this work are as follows:

\begin{itemize}
    \item \textbf{Multi-client heterogeneous training framework}: We propose two training strategies—Sequential and Averaging—that enable multiple clients with different split points to train a shared neural network collaboratively. Sequential strategy processes clients one by one with a shared server model, while the Averaging strategy allows parallel client training with periodic cross-layer parameter aggregation.

    \item \textbf{Server-side cross-layer aggregation mechanism}: We adapt cross-layer aggregation~\cite{Samikwa2024-tg} to the server side to enhance parallel training execution. Our aggregation strategy unifies parameters across different client architectures by averaging common layers from client-specific server models, enabling knowledge transfer between heterogeneous clients despite their different split points.
\end{itemize}

The remainder of this paper is organized as follows. Section II reviews related work in FL, SL, and Early Exit networks. Section III presents our proposed Hetero-SplitEE framework, including detailed descriptions of the Sequential and Averaging training strategies, as well as the architecture details and implementation specifics. Section IV presents our experimental setup and results. Finally, Section V concludes the paper and outlines future research directions.

\section{Related Work}
\textbf{Federated Learning} (FL) is a distributed machine learning paradigm that enables deep neural network training by multiple clients~\cite{McMahan2017-pc}. Each client first trains the deep neural network using its local data and then sends the trained model to the central server, where it aggregates local models from clients to obtain the global model, which encompasses holistic knowledge from all clients.

One of the drawbacks of FL is that all clients have to own and train the same neural network architecture. This feature prevents heterogeneous IoT devices from joining FL, as the varying computational capabilities in FL result in a slowdown in the training process. Since most FL methods are performed synchronously, clients with higher computational capabilities have to wait for ``stragglers,'' devices with lower computational capabilities~\cite{Chen2019-gg, Park2021-zz, Schlegel2023-hw}.

\textbf{Split Learning} (SL) is a distributed machine learning paradigm where a single neural network is split into multiple sub-networks. These different parts are distributed to each client for collaborative training and inference ~\cite{Otkrist2018-av, Vepakomma2018-zg, Samikwa2022-sc}.

When input data is provided, the client that has the first sub-network feeds it to the network to generate an intermediate representation. This representation is then sent to the next sub-network on another client. The data is processed on clients until it reaches the final layer of the complete network.

For model training, backpropagation is performed in the reverse order of the forward pass, propagating gradients back through each sub-network from the final client to the first client.
Recent work has also explored applying SL to increasingly large models, combining it with parameter-efficient training techniques to manage computational and memory constraints~\cite{Lin2024-cm, Zhang2024-te, Ma2025-au}.

\textbf{Early Exit} is a technique that accelerates deep learning inference by allowing a model to output a prediction from an intermediate layer, rather than constantly processing an input through the entire network~\cite{Teerapittayanon2017-xe, Laskaridis2021-bh, Sepehri2023-ip, Chen2023-hp, Zeng2024-se, Samikwa2022-cb, Bajpai2023-ea, Bajpai2024-vp}. This approach is based on the principle that not all inputs require the network's full computational depth to be classified correctly.

Continuing computation after the network reaches a correct output, known as ``overthinking,'' is not only wasteful in terms of time and energy but also can degrade performance by causing the model to change a correct prediction into an incorrect one~\cite{Kaya2018-lq}. To avoid overthinking, early exit augments side branches or internal classifiers (ICs) to the network at various intermediate points. When one of these ICs predicts with sufficient confidence, the model bypasses the remaining layers and immediately outputs that result, saving unnecessary computation.

\section{Proposed Method: Hetero-SplitEE}
\subsection{Architecture Overview}

\begin{figure}[tb]
\centering
\includegraphics[width = 0.8\linewidth]{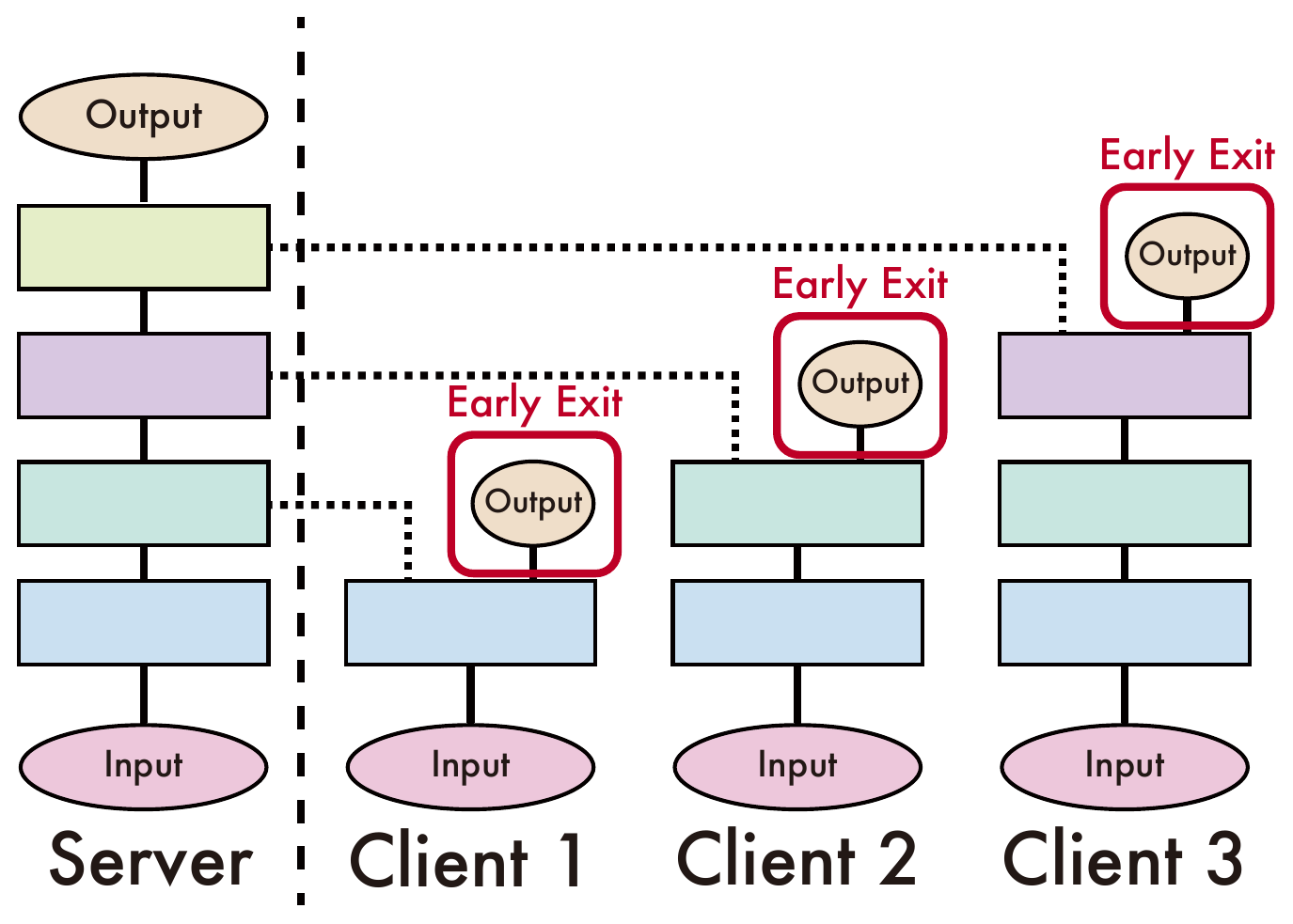}
\caption{Overview of the proposed architecture.}
\label{fig:overview}
\end{figure}

Hetero-SplitEE enables $N$ heterogeneous clients to collaboratively train a shared deep neural network while accommodating their diverse computational capabilities.
The key feature is that, as shown in Fig.~\ref{fig:overview}, different clients can select different split points (cut layers) within the same base architecture.
This feature allows resource-constrained devices to train shallow sub-networks while more capable devices train deeper portions.

In this part, we elaborate on how the proposed architecture processes an input.
Before model training, client $i$ $(i \in [N])$ selects an \textit{end layer} $l_i \in \{1, \ldots, L\}$ based on its computational capacity,
where $L$ is the total number of layers in the base network. 
The client-side network $f_i^{(c)}$ processes input $x$ through layers $1$ to $l_i$, producing intermediate features $h_i = f_i^{(c)}(x)$.
Each client maintains a lightweight \textit{client output layer} $f_i^{(o)}$ attached after layer $l_i$ as an early exit branch.
This output layer serves two purposes: it enables local prediction $\hat{y}_i^{(c)} = f_i^{(o)}(h_i)$ without sending data to the server during inference, and it allows the client to train its sub-network $f_i^{(c)}$ by the loss calculated by local predictions, eliminating the need for synchronous gradient communication during training.
The server-side network processes the remaining layers from $l_i+1$ to $L$, producing the final prediction using the transmitted features $h_i$.
The specific server architecture (shared $f^{(s)}$ or client-specific $f_i^{(s)}$) depends on the training strategy, as described in Section III-B.
The server trains its network using its own loss computed from the final prediction, and importantly, gradients from server training do not propagate back to the client. This design choice prioritizes communication efficiency and enables parallel training execution, as clients can update their parameters independently without waiting for server-side gradients.

\subsection{Training}
We propose two training strategies for the Hetero-SplitEE framework: \textit{Sequential} and \textit{Averaging} strategies.
Both strategies enable heterogeneous clients with different end layers $l_i$ to train a shared neural network collaboratively. Still, they differ fundamentally in how the server-side model is updated and synchronized across clients.
We detail these two strategies in the following sections.

\subsubsection{Sequential Training Strategy}
In the Sequential strategy, all clients share a single server-side network $f^{(s)}$.
While clients execute their training iterations in parallel, the server processes received features sequentially in the order of arrival, updating the shared model progressively with each client's data.
During the global round $t$, all the clients execute their training iteration in parallel.
Client $i$ first performs a forward pass through its client-side network $f_i^{(c)}$ to produce intermediate features $h_i = f_i^{(c)}(x)$, then computes the client-side loss $\mathcal{L}_i^{(c)} = \mathcal{L}(f_i^{(o)}(h_i), y)$ using its output layer $f_i^{(o)}$ and ground truth label $y$.
The client backpropagates this loss to update the client-side network parameters $\theta_i^{(c)}$ without requiring server communication.
Simultaneously, client \(i\) transmits the intermediate features $h_i$ to the server.
The server processes the received $h_i$ in the order of arrival, performing a forward pass through its shared server-side network $f^{(s)}$ to produce the final prediction $\hat{y}_i^{(s)} = f^{(s)}(h_i)$.
The server computes the server-side loss $\mathcal{L}_i^{(s)} = \mathcal{L}(\hat{y}_i^{(s)}, y)$ and updates the shared server model $f^{(s)}$ using this loss.
Note that the server's gradient does not propagate back to the client, maintaining hierarchical training independence.
Round $t$ finishes when the server completes processing all the intermediate features from the clients.

The key advantage of the Sequential strategy is its simplicity and memory efficiency: only one server-side model needs to be maintained throughout training.
However, the single shared server model must process communication from all clients sequentially, which increases total training time and becomes a bottleneck in situations involving numerous clients.
The detailed procedure is shown in Algorithm~\ref{alg:sequential}.

\begin{algorithm}[tb]
\caption{Sequential Training Strategy}
\label{alg:sequential}
\begin{algorithmic}[1]
\REQUIRE Number of clients $N$, end layers $\{l_1, \ldots, l_N\}$, global rounds $T$, local epochs $E$
\ENSURE Trained client-side networks $\{f_1^{(c)}, \ldots, f_N^{(c)}\}$, client output layers $\{f_1^{(o)}, \ldots, f_N^{(o)}\}$, and shared server-side network $f^{(s)}$
\STATE Initialize all networks from the same random seed
\FOR{$t = 1$ to $T$}
    \FOR{$i = 1$ to $N$}
        \STATE \textbf{// Client-side training}
        \FOR{$e = 1$ to $E$}
            \STATE Sample mini-batch $(x, y)$ from client $i$'s local data
            \STATE Compute intermediate features: $h_i = f_i^{(c)}(x)$
            \STATE Compute client prediction: $\hat{y}_i^{(c)} = f_i^{(o)}(h_i)$
            \STATE Compute client loss: $\mathcal{L}_i^{(c)} = \mathcal{L}(\hat{y}_i^{(c)}, y)$
            \STATE Update client networks: $\theta_i^{(c)} \leftarrow \theta_i^{(c)} - \eta \nabla_{\theta_i^{(c)}} \mathcal{L}_i^{(c)}$
            \STATE Update output layer: $\theta_i^{(o)} \leftarrow \theta_i^{(o)} - \eta \nabla_{\theta_i^{(o)}} \mathcal{L}_i^{(c)}$
            \STATE \textbf{// Server-side training (parallel with client)}
            \STATE Send $h_i$ to server
            \STATE Server computes: $\hat{y}_i^{(s)} = f^{(s)}(h_i)$
            \STATE Server computes loss: $\mathcal{L}_i^{(s)} = \mathcal{L}(\hat{y}_i^{(s)}, y)$
            \STATE Server updates: $\theta^{(s)} \leftarrow \theta^{(s)} - \eta \nabla_{\theta^{(s)}} \mathcal{L}_i^{(s)}$
        \ENDFOR
    \ENDFOR
\ENDFOR
\RETURN $\{f_1^{(c)}, \ldots, f_N^{(c)}\}, \{f_1^{(o)}, \ldots, f_N^{(o)}\}, f^{(s)}$
\end{algorithmic}
\end{algorithm}

\subsubsection{Averaging Training Strategy}
The averaging strategy extends the framework to mitigate the bottleneck caused by many clients.
Averaging strategy maintains client-specific server-side networks $f_i^{(s)}$ to process data from clients in parallel, and these networks are periodically synchronized through cross-layer aggregation in the way inspired by FedAvg~\cite{McMahan2017-pc}.
The aggregation method follows the concept of FedAvg, but extends it to heterogeneous networks, as described below.

During each global round $t$, all clients can train in parallel with their corresponding server-side networks.
For each client $i$, the training procedure is similar to the Sequential strategy: the client performs local forward and backward passes using $f_i^{(c)}$ and $f_i^{(o)}$, while the server trains the client-specific model $f_i^{(s)}$ using the transmitted features $h_i$.
The critical difference lies in the aggregation step performed at the end of each round.

Since different clients have different end layers $l_i$, their server-side networks $f_i^{(s)}$ contain distinct subsets of the whole network's layers (from layer $l_i+1$ to layer $L$).
To enable knowledge sharing across heterogeneous architectures, we perform \textit{cross-layer aggregation}~\cite{Samikwa2024-tg} on the server side to enhance parallel training execution: for each layer $l$ in the whole network, we identify all clients whose server-side network contains that layer (i.e., clients with $l_i < l$), forming a set $\mathcal{C}_l = \{i \mid l_i < l\}$.
The parameters of layer $l$ are then averaged across all clients in $\mathcal{C}_l$:
\begin{equation}
    \bar{\theta}_l^{(t)} = \frac{1}{|\mathcal{C}_l|} \sum_{i \in \mathcal{C}_l} \theta_{l,i}^{(t)},
\end{equation}
where $\theta_{l,i}^{(t)}$ represents the parameters of layer $l$ of the server-side network for client \(i\) at round $t$.
These averaged parameters $\bar{\theta}_l^{(t)}$ are then broadcast back to all clients in $\mathcal{C}_l$, synchronizing the common layers across all client-specific server models.
Note that, to enable adequate knowledge sharing across heterogeneous architectures, all models are initialized from the same random seed, ensuring that corresponding layers across different clients start with identical weights.
% This unified initialization is crucial for the cross-layer parameter aggregation performed in our Averaging training strategy.

This cross-layer aggregation mechanism is crucial for enabling effective collaborative learning in heterogeneous environments.
Deeper layers in the network (closer to the output) are shared by more clients, benefiting from aggregated gradients computed from diverse data distributions.
Conversely, shallower layers (closer to the input, immediately after the split point) are trained to optimize classification based on these enriched deeper layers, which is expected to ensure accuracy for intermediate features from clients with shallow end layers.
The aggregation ensures that all clients contribute to and benefit from the shared knowledge encoded in the common layers, regardless of which portions of the network they train.

The averaging strategy enables parallel training, potentially reducing wall-clock training time compared to the Sequential strategy, at the cost of maintaining multiple server-side models and requiring periodic synchronization.
The detailed procedure is shown in Algorithm~\ref{alg:averaged}.

\begin{algorithm}[tb]
\caption{Averaging Training Strategy}
\label{alg:averaged}
\begin{algorithmic}[1]
\REQUIRE Number of clients $N$, end layers $\{l_1, \ldots, l_N\}$, global rounds $T$, local epochs $E$
\ENSURE Trained client-side networks $\{f_1^{(c)}, \ldots, f_N^{(c)}\}$, client output layers $\{f_1^{(o)}, \ldots, f_N^{(o)}\}$, and client-specific server-side networks $\{f_1^{(s)}, \ldots, f_N^{(s)}\}$
\STATE Initialize all networks from the same random seed
\FOR{$t = 1$ to $T$}
    \STATE \textbf{// Parallel client and server training}
    \FOR{$i = 1$ to $N$ \textbf{in parallel}}
        \FOR{$e = 1$ to $E$}
            \STATE Sample mini-batch $(x, y)$ from client $i$'s local data
            \STATE \textbf{// Client-side training}
            \STATE Compute intermediate features: $h_i = f_i^{(c)}(x)$
            \STATE Compute client prediction: $\hat{y}_i^{(c)} = f_i^{(o)}(h_i)$
            \STATE Compute client loss: $\mathcal{L}_i^{(c)} = \mathcal{L}(\hat{y}_i^{(c)}, y)$
            \STATE Update client networks: $\theta_i^{(c)} \leftarrow \theta_i^{(c)} - \eta \nabla_{\theta_i^{(c)}} \mathcal{L}_i^{(c)}$
            \STATE Update output layer: $\theta_i^{(o)} \leftarrow \theta_i^{(o)} - \eta \nabla_{\theta_i^{(o)}} \mathcal{L}_i^{(c)}$
            \STATE \textbf{// Server-side training (parallel with client)}
            \STATE Send $h_i$ to server
            \STATE Server computes: $\hat{y}_i^{(s)} = f_i^{(s)}(h_i)$
            \STATE Server computes loss: $\mathcal{L}_i^{(s)} = \mathcal{L}(\hat{y}_i^{(s)}, y)$
            \STATE Server updates: $\theta_i^{(s)} \leftarrow \theta_i^{(s)} - \eta \nabla_{\theta_i^{(s)}} \mathcal{L}_i^{(s)}$
        \ENDFOR
    \ENDFOR
    \STATE \textbf{// Cross-layer aggregation}
    \FOR{each layer $l$ in the full network}
        \STATE Identify clients with layer $l$: $\mathcal{C}_l = \{i \mid l_i < l\}$
        \IF{$|\mathcal{C}_l| > 0$}
            \STATE Averaging parameters: $\bar{\theta}_l^{(t)} = \frac{1}{|\mathcal{C}_l|} \sum_{i \in \mathcal{C}_l}\theta_{l,i}^{(t)}$
            \STATE Broadcast $\bar{\theta}_l^{(t)}$ to all clients in $\mathcal{C}_l$
            \FOR{$i \in \mathcal{C}_l$}
                \STATE Update layer $l$ in $f_i^{(s)}$: $\theta_{l,i}^{(t)} \leftarrow \bar{\theta}_l^{(t)}$
            \ENDFOR
        \ENDIF
    \ENDFOR
\ENDFOR
% \RETURN $\{f_1^{(c)}, \ldots, f_N^{(c)}\}, \{f_1^{(o)}, \ldots, f_N^{(o)}\}, \{f_1^{(s)}, \ldots, f_N^{(s)}\}$
\RETURN \resizebox{0.85\linewidth}{!}{$
    \{f_1^{(c)}, \ldots, f_N^{(c)}\},
    \{f_1^{(o)}, \ldots, f_N^{(o)}\},
    \{f_1^{(s)}, \ldots, f_N^{(s)}\}
$}
\end{algorithmic}
\end{algorithm}

\subsection{Inference}
Each client collaborates with the server to perform inference using an entropy-based confidence mechanism.
For a given input, the client feeds it to its local network to obtain both an early-exit prediction from the internal classifier and an intermediate representation from the sub-network.
If the internal classifier's prediction confidence exceeds a predefined threshold, the early-exit output is used as the final prediction, avoiding communication overhead and server-side computation.
In contrast, if the confidence is below the threshold, the intermediate representation is transmitted to the server for deeper processing, with the server-side network's output serving as the final prediction.

This adaptive inference framework offers multiple benefits: (1) it accelerates inference by enabling early exits for simple instances that do not require deep processing, (2) it reduces communication overhead by minimizing data transmission to the server, and (3) it efficiently allocates computational resources between clients and servers based on input complexity and available client capabilities.

During inference, each client cooperatively works with the server using an entropy-based early exit mechanism (Algorithm~\ref{alg:entropy_inference}).
For input $x_b$, client $i$ computes intermediate features $h_i = f_i^{(c)}(x_b)$ and local prediction $\hat{y}_i^{(c)} = f_i^{(o)}(h_i)$.
The prediction confidence is measured using entropy: $H_i^{(c)} = -\sum_j p_{i,j}^{(c)} \log p_{i,j}^{(c)}$, where $p_i^{(c)} = \text{softmax}(\hat{y}_i^{(c)})$.
Lower entropy indicates higher confidence.

Client $i$ compares its confidence score $C_i^{(c)} = -H_i^{(c)}$ against threshold $\tau$.
If $C_i^{(c)} > \tau$, the client's prediction is used directly ($\hat{y}_{\text{final}} = \arg\max(p_i^{(c)})$), achieving early exit.
Otherwise, $h_i$ is sent to the server for deeper processing, and $\hat{y}_{\text{final}} = \arg\max(\text{softmax}(f^{(s)}(h_i)))$.

The threshold $\tau$ balances accuracy and efficiency: lower values encourage early exits, reducing communication and latency but potentially sacrificing accuracy. In comparison, higher values improve accuracy at the cost of increased server load.
We analyze this trade-off by evaluating thresholds in the range $[0.0, 4.0]$ with a granularity of 0.05, measuring accuracy, client/server selection rates, and average entropy.
This framework enables adaptive inference where simple inputs exit early at the client. In contrast, complex inputs benefit from server-side deep processing, which reduces communication overhead and improves scalability in heterogeneous IoT environments.

\begin{figure}[ht]
    \begin{algorithm}[H]
        \caption{Entropy-based Early Exit Inference Algorithm}
        \label{alg:entropy_inference}
        \begin{algorithmic}[1]
            \STATE \textbf{Input:} Image $(x_b)$, confidence threshold $\tau$
            \STATE \textbf{Output:} Final prediction $\hat{y}_{final}$
            
            \STATE \textbf{Phase 1: Client-side inference}
            \STATE $h_i = f_i^{(c)}(x_b; \theta_i^{(c)})$ \COMMENT{Forward to cut layer $l_i$}
            \STATE $\hat{y}_i^{(c)} = f_i^{(o)}(h_i; \theta_i^{(o)})$ \COMMENT{Client prediction}
            \STATE $p_i^{(c)} = \text{softmax}(\hat{y}_i^{(c)})$ \COMMENT{Client probabilities}
            \STATE $H_i^{(c)} = -\sum p_i^{(c)} \log p_i^{(c)}$ \COMMENT{Client entropy}
            \STATE $C_i^{(c)} = -H_i^{(c)}$ \COMMENT{Client confidence}
            
            \STATE \textbf{Phase 2: Confidence-based decision}
            \IF{$C_i^{(c)} > \tau$}
                \STATE $\hat{y}_{final} = \arg\max(p_i^{(c)})$ \COMMENT{Early exit with client prediction}
            \ELSE
                \STATE \textbf{Phase 3: Server-side inference}
                \STATE Send $h_i$ to server \COMMENT{Feature transmission}
                \STATE $\hat{y}_i^{(s)} = f^{(s)}(h_i; \theta^{(s)})$ \COMMENT{Server prediction}
                \STATE $\hat{y}_{final} = \arg\max(\text{softmax}(\hat{y}_i^{(s)}))$ \COMMENT{Use server prediction}
            \ENDIF
            
            \STATE \textbf{return} $\hat{y}_{final}$
        \end{algorithmic}
    \end{algorithm}
\end{figure}

\section{Experiments}

\subsection{Experimental Setup}
\subsubsection{Datasets}
Split neural networks are trained using the proposed framework for image classification tasks using CIFAR-10, CIFAR-100, and STL-10. CIFAR-10 and CIFAR-100 consist of 50,000 training images and 10,000 test images of size \(32\times32\), respectively. STL-10 contains 5,000 training images and 8,000 test images of size \(96\times96\). During training, all the images are normalized and augmented.
In augmentation, we zero-pad images by 4 pixels on all sides, take a random crop of the original size, and randomly flip the pictures.
For inference, images are not augmented; instead, they are normalized.
For multi-client training, datasets are split uniformly at random and distributed to clients, resulting in independent and identically distributed (IID) datasets within each client.

\subsubsection{Models}
We use ResNet-18 as a base neural network. The network is split into two sub-networks, the client-side network and the server-side network (Table~\ref{tab:resnet18_client_architecture}).
\begin{table*}[t]
    
\centering
\caption{ResNet18 Based Model Architecture for Experiments}
\label{tab:resnet18_client_architecture}
\scalebox{0.9}{
\begin{tabular}{|c|c|c|c|c|}
\hline
\textbf{Layer Name} & \textbf{Components} & \begin{tabular}{c}  \textbf{Output}\\ \textbf{Channels} \end{tabular} & \textbf{Stride} & \textbf{Notes} \\
\hline
\hline
\multicolumn{5}{|c|}{\textbf{Input Layer}} \\
\hline
Input & - & 3 & - & \begin{tabular}{c} 32×32×3 (CIFAR-10/100)\\96×96×3 (STL-10)\end{tabular} \\
\hline
\hline
\multicolumn{5}{|c|}{\textbf{Main Layers}} \\
\hline
Layer1 & \begin{tabular}{c|c} 
\textbf{CIFAR10/100} & \textbf{Other} \\
\hline
Conv2d(3×3, padding=1) & Conv2d(7×7, padding=3) \\
BatchNorm2d & BatchNorm2d \\
ReLU & ReLU \\
Identity & MaxPool2d(3×3, stride=2)
\end{tabular} & 64 & \begin{tabular}{c} 1 (CIFAR) \\ 2 (Other) \end{tabular} & \begin{tabular}{c} Dataset-specific configuration \\ No MaxPool for CIFAR \\ Included if end\_layer $\ge$ 1 \end{tabular} \\
\hline
Layer2 & BasicBlock × 2 & 64 & 1 & \begin{tabular}{c} Included if end\_layer  $\ge$ 2 \end{tabular} \\
\hline
Layer3 & BasicBlock × 2 & 64 & 1 & \begin{tabular}{c} Included if end\_layer  $\ge$ 3 \end{tabular} \\
\hline
Layer4 & BasicBlock × 2 & 128 & 2 & \begin{tabular}{c}Included if end\_layer $\ge$ 4 \end{tabular} \\
\hline
Layer5 & BasicBlock × 2 & 256 & 2 & \begin{tabular}{c} Included if end\_layer  $\ge$ 5 \end{tabular} \\
\hline
Layer6 & BasicBlock × 2 & 512 & 2 & \begin{tabular}{c}  Included if end\_layer  $\ge$ 6\end{tabular} \\
\hline
\hline
\multicolumn{5}{|c|}{\textbf{Output Layers}} \\
\hline
 \begin{tabular}{c} Server\\OutputLayer\end{tabular} & \begin{tabular}{c} AdaptiveAvgPool2d(1,1) \\ Flatten \\ Linear \end{tabular} & num\_classes & - & \begin{tabular}{c} Use for Server inference\end{tabular} \\
\hline
\begin{tabular}{c} Client\\OutputLayer\end{tabular} & \begin{tabular}{c} AdaptiveAvgPool2d(1,1) \\ Flatten \\ Linear \end{tabular} & num\_classes & - & \begin{tabular}{c} Use for Client inference \\ Input channels depend on end\_layer \end{tabular} \\
\hline
\end{tabular}
}
\end{table*}
The client-side network has a side branch after the last layer of the sub-network, allowing for early exit of inference.
The side branch is composed of an average pooling layer and a fully connected layer.
All the models start with the same initial weights and biases.

\subsubsection{Hyperparameters}
The hyperparameters used for model training are shown in Table~\ref{tab:params}.
The learning rate is decayed every epoch by cosine annealing from the initial learning rate \(lr_0 = 0.001\) to the minimum learning rate \(\eta_{\text{min}} = 1.0\times10^{-6}\) (corresponding to \(lr_0/1000\)).
We use the Adam optimizer~\cite{Kingma2014-af}.
All model parameters, including client-side models, server-side models, and output layers, are optimized using the same hyperparameters, except for the server learning rate in the Sequential strategy, which is divided by 12.0 to account for sequential updates from multiple clients.

\begin{table}[tb]
    \centering
    \caption{Hyperparameters for model training}
    \label{tab:params}
    \begin{tabular}{@{}ll@{}}\toprule
        Epochs & 600 \\
        Optimizer & Adam \\
        Scheduler & Cosine Annealing \\
        \quad Warmup Epochs & 0 \\
        \quad $T_{\text{max}}$ & 600 \\
        \quad $\eta_{\text{max}}$  & 0.001 \\
        \quad\quad (Sequential strategy server) & $0.001 / 12.0 = 8.33\times10^{-5}$ \\
        \quad $\eta_{\text{min}}$ & $1.0\times10^{-6}$ \\
        Batch Size & 1024 (CIFAR), 64 (STL) \\
        \bottomrule
    \end{tabular}
\end{table}
\subsubsection{Training Settings}
We evaluate our proposed framework under various experimental conditions to demonstrate its effectiveness and flexibility.
All experiments involve 12 clients with a 100\% participation rate (i.e., all clients participate in each training round).

\paragraph{Homogeneous setting}
In the homogeneous setting, all 12 clients use the same end layer position.
We conduct experiments for three different end layer configurations: end\_layer=3, end\_layer=4, and end\_layer=5, where each configuration is trained separately.
This setting serves as a controlled baseline to isolate the effects of client heterogeneity and validate that our framework also functions effectively when all clients have identical computational constraints.

\paragraph{Heterogeneous setting}
In the heterogeneous setting, clients have different end-layer positions to simulate environments with diverse computational capabilities.
Specifically, among the 12 clients, 4 clients use end\_layer=3, 4 clients use end\_layer=4, and 4 clients use end\_layer=5.
This configuration represents realistic scenarios where edge devices have varying resource constraints.

\paragraph{Baseline comparisons}
We compare our proposed framework against two baseline approaches:

\textbf{Centralized baseline:} All training data from all clients are aggregated at a central location, and a single model using our proposed architecture is trained on the combined dataset.
This represents the upper bound of performance achievable with centralized training, without constraints on privacy or communication.

\textbf{Distributed baseline:} Each client independently trains its own model using our proposed architecture on its local data without any server-side information aggregation or model sharing.
This represents the lower bound where clients do not benefit from collaborative learning, serving as a comparison to demonstrate the effectiveness of training with server-side aggregation.

All settings, including the distributed one, are trained with identical data partitions and hyperparameters so that observed performance differences isolate the effect of collaborative aggregation mechanisms.

\subsection{Training with Homogeneous Client Models}
Table~\ref{tab:result-homo-without-fc-fixed} shows the classification accuracy results for homogeneous client configurations across three datasets (CIFAR-10, CIFAR-100, and STL-10) and three end layer positions (Layer-3, Layer-4, and Layer-5).
In this setting, all 12 clients use the same end-layer position in each experiment.

We compare two training strategies of our proposed framework: Algorithm 1 (Sequential strategy) and Algorithm 2 (Averaging strategy), against two baselines: Centralized and Distributed training.
The results show both server-side outputs and client-side early-exit outputs.

The Centralized baseline achieves significantly higher accuracy than all other approaches, demonstrating the performance upper bound when all data is aggregated centrally without privacy or communication constraints.

An important observation is that the effectiveness of our proposed collaborative learning framework varies with task difficulty.
For CIFAR-10, which is a relatively easy 10-class classification task with high overall accuracy (67--73\%), our proposed methods (Algorithms 1 and 2) show similar performance to the Distributed baseline, indicating limited benefits from server-side aggregation for simpler tasks.

However, as task complexity increases, the advantages of our framework become apparent.
For CIFAR-100, a challenging 100-class classification task with lower overall accuracy (25--39\%), our proposed methods demonstrate notable improvements over the Distributed baseline, particularly for server-side outputs.
For example, with Layer-3 configuration, Algorithm 1 achieves 39.37\% server-side accuracy, compared to 25.81\% for Distributed training, demonstrating the significant value of collaborative learning for complex tasks.

STL-10 results show intermediate characteristics, with modest improvements over distributed training, consistent with its moderate task difficulty.

A limitation of the proposed method is that it does not incorporate a mechanism to align the representations from different clients. Consequently, other clients may generate dissimilar intermediate representations even from the same input. While we hypothesize that aligning these representations could further enhance performance, we leave this investigation for future work.

\begin{table*}[t]
    \centering
    \caption{Results with homo client models (without-conv)}
    \begin{tabular}{@{}lllllllllll} \toprule
                    &        & \multicolumn{3}{c}{CIFAR10} & \multicolumn{3}{c}{CIFAR100} & \multicolumn{3}{c}{STL-10} \\ \cmidrule(lr){3-5}\cmidrule(lr){6-8}\cmidrule(lr){9-11}
             Method & Location & Layer-3 & Layer-4 & Layer-5 & Layer-3 & Layer-4 &  Layer-5 & Layer-3 & Layer-4 & Layer-5 \\ \midrule
        Sequential strategy &   Server &   72.35 &   71.53 &   73.71 &   39.37 &   34.73 &   33.69 &   50.16 &   49.79 &   47.61  \\                   (Ours; Alg. 1) &   Client &   68.21 &   70.39 &   73.59 &   28.52 &   30.46 &   33.77 &   48.96 &   49.10 &   48.11  \\ \midrule
        Averaging strategy &   Server &   68.83 &   70.68 &   73.71 &   37.88 &   35.37 &   33.73 &   50.23 &   48.99 &   47.36  \\                   (Ours; Alg. 2) &   Client &   67.79 &   70.36 &   73.82 &   28.20 &   30.70 &   33.55 &   49.17 &   48.42 &   47.59  \\ \midrule
        Centralized &   Server &   86.99 &   89.96 &   93.16 &   60.33 &   66.01 &   70.03 &   75.08 &   76.41 &   75.72  \\                    &   Client &   86.59 &   89.98 &   93.23 &   55.19 &   64.16 &   70.49 &   74.79 &   76.25 &   75.81  \\ \midrule
        Distributed &   Server &   68.35 &   70.75 &   73.44 &   25.81 &   28.18 &   31.82 &   48.62 &   47.92 &   46.51  \\                   (Baseline) &   Client &   67.81 &   70.36 &   73.82 &   28.32 &   30.70 &   33.90 &   49.17 &   48.86 &   46.98 \\\bottomrule
    \end{tabular}
    \label{tab:result-homo-without-fc-fixed}
\end{table*}

\begin{table*}[t]
    \centering
    \caption{Results with hetero client models (without-conv)}
    \begin{tabular}{@{}lllllllllll} \toprule
                    &        & \multicolumn{3}{c}{CIFAR10} & \multicolumn{3}{c}{CIFAR100} & \multicolumn{3}{c}{STL-10} \\ \cmidrule(lr){3-5}\cmidrule(lr){6-8}\cmidrule(lr){9-11}
             Method & Location & Layer-3 & Layer-4 & Layer-5 & Layer-3 & Layer-4 &  Layer-5 & Layer-3 & Layer-4 & Layer-5 \\ \midrule
        Sequential strategy &   Server &   68.19 &   70.35 &   74.16 &   33.49 &   33.10 &   34.12 &   50.49 &   47.79 &   47.48  \\                    (Ours; Alg. 1) &   Client &   67.44 &   70.44 &   74.43 &   28.51 &   28.45 &   34.63 &   50.56 &   47.11 &   46.96  \\ \midrule
        Averaging strategy &   Server &   68.01 &   70.41 &   74.61 &   34.93 &   33.37 &   35.05 &   51.01 &   46.73 &   46.70  \\                    (Ours; Alg. 2) &   Client &   67.74 &   70.38 &   74.31 &   27.91 &   30.07 &   35.29 &   50.23 &   46.58 &   47.26  \\ \midrule
        Centralized &   Server &   86.99 &   89.96 &   93.16 &   60.33 &   66.01 &   70.03 &   75.08 &   76.41 &   75.72  \\                    &   Client &   86.59 &   89.98 &   93.23 &   55.19 &   64.16 &   70.49 &   74.79 &   76.25 &   75.81  \\ \midrule
        Distributed &   Server &   68.40 &   69.90 &   73.53 &   24.82 &   26.53 &   32.05 &   48.91 &   45.95 &   46.85  \\                   (Baseline)  &   Client &   67.29 &   69.99 &   73.62 &   27.58 &   28.69 &   33.91 &   50.23 &   46.58 &   47.26 \\\bottomrule
    \end{tabular}
    \label{tab:result-hetero-without-fc-fixed}
\end{table*}

\subsection{Training with Heterogeneous Client Models}
Table~\ref{tab:result-hetero-without-fc-fixed} presents results for heterogeneous client configurations, where clients have different end layer positions to simulate diverse computational capabilities.
Specifically, among the 12 clients, 4 clients use end\_layer=3, 4 clients use end\_layer=4, and 4 clients use end\_layer=5.
The table reports accuracy for each end layer depth separately, showing how different client types perform within the same federated training session.

Note that an accurate Centralized baseline cannot be realized in the heterogeneous setting, since it does not naturally accommodate multiple end-layer positions.

For reference, Centralized results from the homogeneous setting (Table~\ref{tab:result-homo-without-fc-fixed}) provide an approximate upper bound.
The task complexity trend observed in the homogeneous setting also persists here: improvements over the Distributed baseline are modest for CIFAR-10 but more pronounced for CIFAR-100, confirming that collaborative learning benefits increase with task difficulty.

Notably, Algorithm 1 (Sequential strategy) and Algorithm 2 (Averaging strategy) achieve more similar performance in the heterogeneous setting compared to the homogeneous case.
In the homogeneous setting, the Sequential strategy benefits from processing intermediate features with consistent dimensions and characteristics across all clients, resembling conventional centralized training.
However, in the heterogeneous setting, this advantage diminishes as the server must handle intermediate features from varying depths with different dimensions and semantic levels, narrowing the performance gap with the Averaging strategy.
Nevertheless, both methods demonstrate superior performance to distributed training on sufficiently challenging tasks, indicating that both approaches possess the capability to integrate information from multiple clients in heterogeneous environments appropriately.

\subsection{Sensitivity Analysis to Early Exiting Thresholds}

In this section, we analyze the sensitivity of collaborative inference to the confidence threshold that determines whether to adopt early exit predictions.

\begin{figure}[ht]
\centering
\includegraphics[width = 0.8\linewidth]{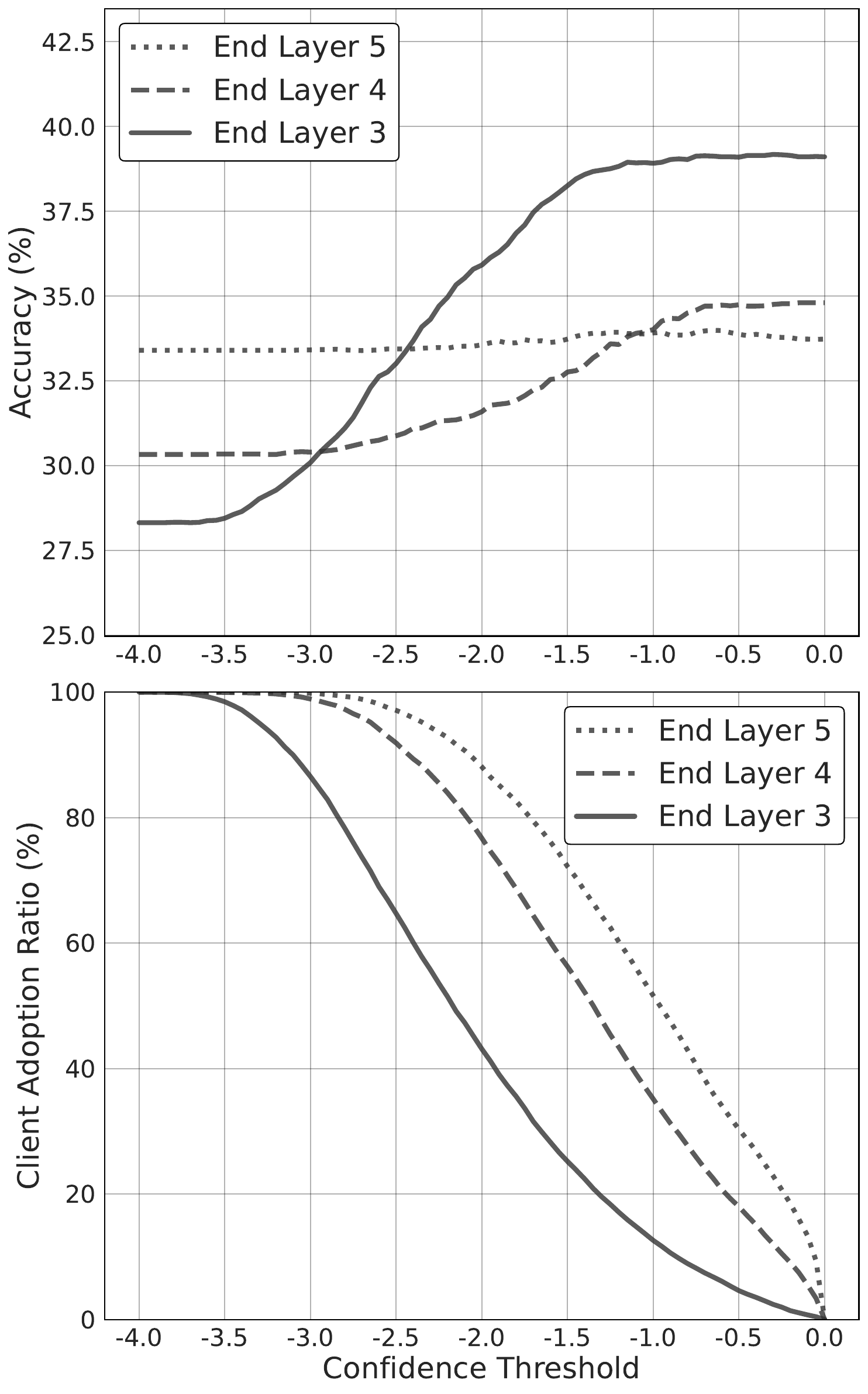}
\caption{(top) Classification accuracy with different confidence thresholds for inference offloading. (bottom) Adoption ratio of early-exit inference on the client side with different confidence thresholds.}
\label{fig:sa}
\end{figure}

Fig.~\ref{fig:sa} shows the classification accuracy (top) and the client adoption ratio (bottom) with different confidence thresholds on CIFAR-100 with homogeneous client models that are trained by the Sequential strategy.
Larger confidence thresholds make the architecture more conservative by rejecting more early-exit inferences and offloading them to the server.
We demonstrated that the more conservative the architecture, the more accurate the predictions are, regardless of the location of the split points.
This is because additional computation on the server side is beneficial for complex tasks that cannot be handled by the client side alone, such as CIFAR-100.

However, conservative architectures with higher thresholds result in slower inference because more inputs are transmitted to the server, requiring more computation and communication cost, as shown at the bottom of Fig.~\ref{fig:sa}.
This feature highlights the importance of setting the confidence threshold appropriately to strike a balance between expected latency for inference and expected accuracy of inference.

\section{Conclusion}
In this work, we proposed Hetero-SplitEE, a framework for collaborative training and inference with split neural networks across multiple clients, which incorporates two distinct training strategies. Our empirical results demonstrated that both proposed strategies are effective in heterogeneous IoT scenarios, each exhibiting unique strengths and weaknesses. Notably, the effectiveness of the proposed method increases with task difficulty, an improvement attributable to its collaborative training and inference capabilities. Furthermore, our analysis highlights the importance of setting an appropriate confidence threshold to strike a balance between inference latency and accuracy. We anticipate that results may vary depending on the specific neural network architecture and task. Therefore, investigating the efficacy of our method in such diverse cases remains a key direction for future work.
Moreover, further investigation is needed on the performance under various settings. This includes scenarios with non-iid data distributions, which are common in real-world IoT deployments, as well as situations involving different architectures, including large language models.

\section*{Acknowledgments}
This work was supported by JST CREST (JPMJCR21D2).

% \printbibliography
\bibliographystyle{IEEEtran}
\bibliography{references}

\end{document}